\documentclass[letterpaper, 10 pt, conference]{format/ieeeconf}

\usepackage{amsmath}
\usepackage{amssymb}
\usepackage{amsfonts}

\usepackage{amsthm}
\usepackage[ruled,vlined]{algorithm2e}
\usepackage{mathtools}
\usepackage{tabularx}
\usepackage{graphicx}
\usepackage{subcaption} 
\usepackage{enumerate}
\usepackage{booktabs}
\usepackage{float}
\usepackage{url}
\usepackage{verbatim}
\usepackage{booktabs} 
\usepackage{multirow}
\usepackage{color}
\usepackage{pifont}
\usepackage[linkcolor=black,citecolor=black,urlcolor=black,colorlinks=true]{hyperref}
\usepackage{cite}
\usepackage{mathrsfs}

\graphicspath{{figures/}}
\IEEEoverridecommandlockouts
\let\oldtwocolumn\twocolumn
\renewcommand\twocolumn[1][]{%
    \oldtwocolumn[{#1}{
    \includegraphics[width=0.99\textwidth]{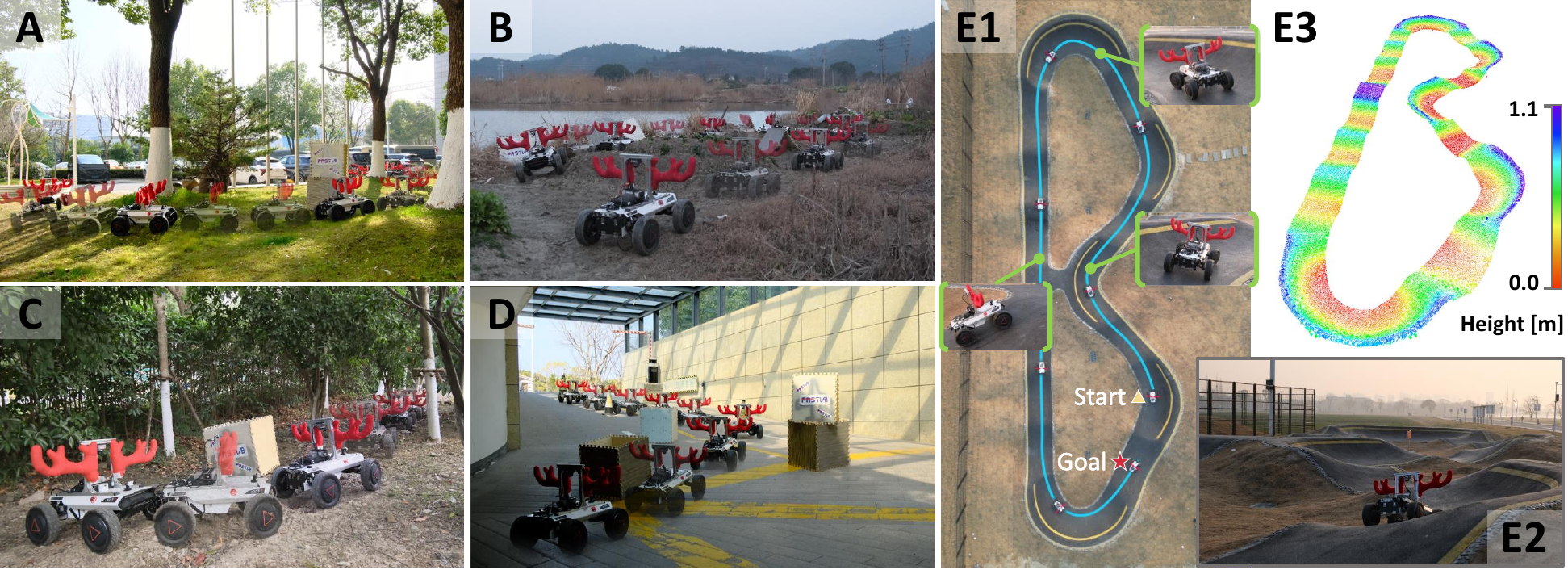}
    \captionof{figure} {SEB-Naver on the grasslands (A), farmland (B), forests (C), underground parking lots (D), and pump track (E).}
    \label{fig:real_exp}
    \vspace{0.4cm}
    }]
}

\title{SEB-Naver: A $SE(2)$-based Local Navigation\\ Framework for Car-like Robots on Uneven Terrain}

\author{Xiaoying Li$^{{\dagger}2}$, Long Xu$^{{\dagger}1,2}$, Xiaolin Huang$^{{\dagger}2}$, Donglai Xue$^{2}$, Zhihao Zhang$^{2}$,\\ Zhichao Han$^{1,2}$, Chao Xu$^{1,2}$, Yanjun Cao$^{2}$, and Fei Gao$^{1,2}$
\thanks{$^{\dag}$Indicates equal contribution.}
\thanks{$^{1}$State Key Laboratory of Industrial Control Technology, Zhejiang University, Hangzhou 310027, China. \textit{Corresponding author: Fei Gao}}
\thanks{$^{2}$Huzhou Institute of Zhejiang University, Huzhou 313000, China.}
\thanks{E-mail: {\tt\small \{gaolon, fgaoaa\}@zju.edu.cn}}
}

\begin{document}

    \maketitle
    \thispagestyle{empty}
    \pagestyle{empty}

\begin{abstract}
Autonomous navigation of car-like robots on uneven terrain poses unique challenges compared to flat terrain, particularly in \textit{traversability assessment} and \textit{terrain-associated kinematic} modelling for motion planning. This paper introduces SEB-Naver, a novel $SE(2)$-based local navigation framework designed to overcome these challenges. First, we propose an efficient traversability assessment method for $SE(2)$ grids, leveraging GPU parallel computing to enable real-time updates and maintenance of local maps. Second, inspired by differential flatness, we present an optimization-based trajectory planning method that integrates terrain-associated kinematic models, significantly improving both planning efficiency and trajectory quality. Finally, we unify these components into SEB-Naver, achieving real-time terrain assessment and trajectory optimization. Extensive simulations and real-world experiments demonstrate the effectiveness and efficiency of our approach. The code is at \url{https://github.com/ZJU-FAST-Lab/seb_naver}.
\end{abstract}

\section{Introduction}
\label{sec:Introduction}
Autonomous navigation of car-like robots on uneven terrain requires additional considerations compared to flat terrain, specifically for  \textit{traversability assessment} and \textit{terrain-associated kinematic model}. The former is an additional module integrated into the autonomous navigation framework, which estimates the risk of anomalies based on the terrain and state of the robot. The latter is a factor that must be considered in the trajectory generation module to ensure that the planned trajectory can be well tracked by the robot.

\begin{figure}[t]
    \centering
    \includegraphics[width=1.0\columnwidth]{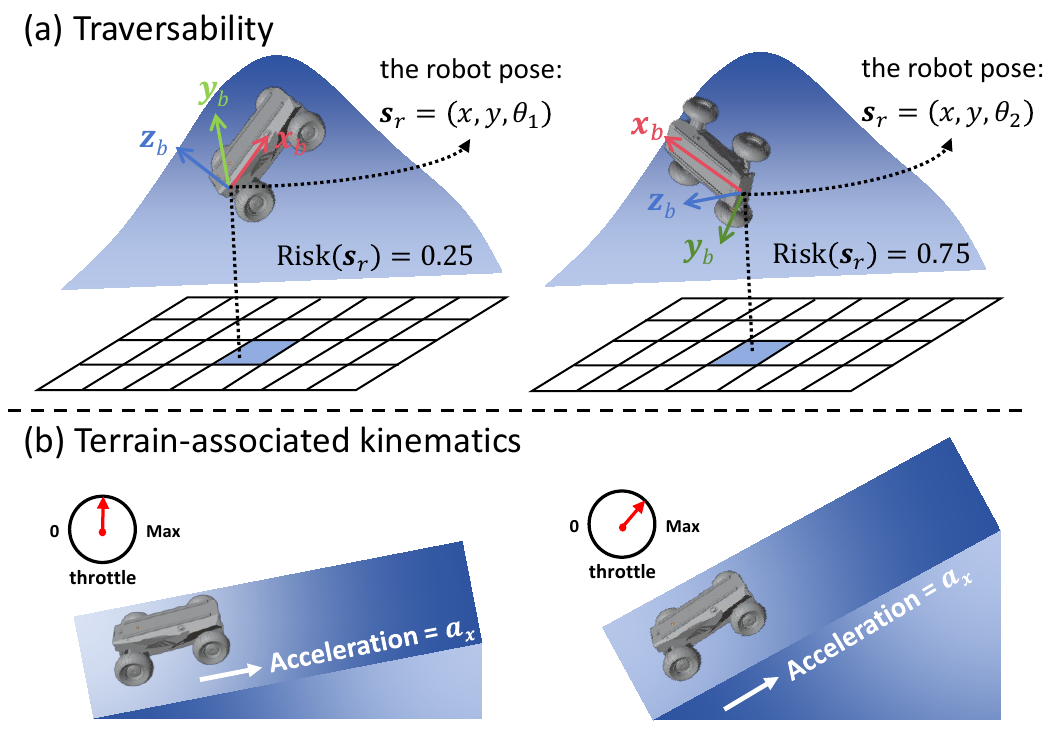}
	\caption{Figure (a) illustrates that different robot poses with different pitch and roll angles may be projected onto the same grid cell in $\mathbb{R}^2$. Figure (b) illustrates that the throttle required for the robot to achieve the same acceleration $a_x$ is different due to the presence of gravity.
 }\label{fig:intro}
\end{figure}

Classical methods\cite{fankhauser2018probabilistic,elegpu,step,elevation} for \textit{traversability assessment} tend to focus only on the geometric information of the terrain, such as slope, curvature, etc. To achieve real-time updates in local navigation frameworks, they usually construct robot-centric grid maps by discretizing the $\mathbb{R}^2$ space, leading to conservative or aggressive results. As shown in Fig.~\ref{fig:intro} (a), different robot poses with varying pitch and roll angles may be projected onto the same grid cell in $\mathbb{R}^2$ space. However, since the tolerance of the robot to lateral tilt and backward tilt differs, the risk values should also vary. A more detailed evaluation of traversability based on the robot state in $SE(3)$ often incurs a significantly higher computational burden\cite{drivingonpoint}. Moreover, due to the substantial increase in spatial dimensions, using discrete grids to achieve real-time updates becomes impractical. 

In our previous work\cite{iros2023}, we propose \textit{terrain pose mapping} to describe the impact of terrain on the robot, allowing us to derive the $SE(3)$ state of the robot from a given state in $SE(2)$. However, the $SE(2)$ space remains too large to meet real-time requirements. In this work, we draw inspiration from parallel computing, proposing an efficient traversability assessment method for $SE(2)$ grids using GPU, achieving real-time updates and maintenance of the local map. 

Fig.~\ref{fig:intro} (b) illustrates how the kinematic model of the robot is influenced by the terrain. Due to gravity, the throttle required to achieve the same acceleration in the body frame varies with the slope. If the planner ignores this factor, the generated trajectory may become infeasible for the robot. Therefore, the \textit{terrain-associated kinematic model} must be considered by the trajectory planner to enable the controller to better track the trajectory. Due to the nonlinearity of the terrain and the high-dimensional robot state space, which increases the dimensionality of the problem, most existing works\cite{travelhybrid,drivingonpoint,putn,iros2023} struggle to effectively incorporate \textit{terrain-associated kinematic model} into planning, making it challenging to balance efficiency and trajectory quality.

Differential flatness has demonstrated its efficiency in trajectory planning for autonomous driving\cite{hzc,unpublished_han}. In this paper, we find that although the presence of uneven terrain complicates the kinematics, the state of the robot can still be represented using flat outputs through \textit{terrain pose mapping}, simplifying the optimization problem. Furthermore, our derivations reveal that the \textit{terrain-associated kinematic} constraints can also be analytically modeled by flat outputs. Based on this, we propose a novel optimization-based trajectory planning method for car-like robots on uneven terrain, significantly improving efficiency and trajectory quality. Moreover, we integrate it with the designed real-time traversability assessment algorithm for $SE(2)$ grids, proposing SEB-Naver, a $SE(2)$-based local navigation framework. Extensive simulations and real-world experiments validate the efficiency and effectiveness of our approach. The contributions of the paper are:

\begin{itemize}
\item [1)] We propose a local mapping pipeline that facilitates real-time traversability assessment for $SE(2)$ grids by utilizing the parallel computing power of GPU.
\item [2)] Based on differential flatness, we propose a novel trajectory optimization model that can effectively consider \textit{terrain-associated kinematics} and significantly improve the efficiency of car-like robots on uneven terrain.
\item [3)] Integrating the above two modules, we propose SEB-Naver, a $SE(2)$-based local navigation framework. Extensive simulations and real-world experiments verify the practicality and efficiency of our approach.
\end{itemize}

\section{Related Works}
\label{sec:RelatedWork}
\subsection{LiDAR-based Traversability Assessment}
\label{subsec:LiDAR-based Traversability Assessment}
Traversability assessment is a critical component for safe and robust navigation on rough terrain. Based on LiDAR, most of the existing works~\cite{fankhauser2018probabilistic,elegpu,step,elevation} primarily focus on the geometric features of the terrain, such as ground height, curvature, slope, etc. To meet real-time requirements, these approaches construct and maintain a robot-centric grid map by discretizing the $\mathbb{R}^2$ space, ensuring efficiency. However, this reliance on geometric features can lead to overly conservative or aggressive terrain evaluation results. For instance, different robot poses with varying pitch and roll angles may be projected onto the same grid cell in $\mathbb{R}^2$ when the evaluation is centred on the rear axle of a car-like robot, as shown in Fig.~\ref{fig:intro} (a). To address the issue of overly conservative or aggressive evaluation results, some works\cite{putn,drivingonpoint,iros2023} estimate traversability in $SE(3)$ space. This allows for a more detailed consideration of pose-related variables, such as the pitch and roll angles of the robot, as well as suspension\cite{suspension}. However, due to the high dimensionality of the $SE(3)$ space (locally homeomorphic to $\mathbb{R}^6$), most methods\cite{putn,drivingonpoint} tend to perform the evaluation when traversability values are required, which may significantly impact the efficiency of trajectory planning. 

In our previous work~\cite{iros2023}, \textit{terrain pose mapping} was proposed to enable the construction of grid maps in the lower-dimensional $SE(2)$ space for traversability assessment, improving the efficiency of trajectory generation. However, the number of grids generated by discretizing $SE(2)$ remains relatively large, making it insufficient to support real-time updates, as realized with local elevation maps~\cite{fankhauser2018probabilistic}. In this paper, we leverage the powerful parallel computing capabilities of GPUs to design a local mapping framework in $SE(2)$ space, achieving real-time performance.

On the other hand, more techniques, such as neural networks\cite{learn1,learn2,learn3,learn4} and the Gaussian process, have recently been applied to traversability assessment. Some works~\cite{learn_self1,learn_self2,learn_self3,learn_self4} employ self-supervised learning methods to compute traversability maps, which have been demonstrated to be effective for autonomous navigation in unstructured outdoor environments. The main limitation of these learning-based approaches lies in the uncertainty of the models, which can significantly affect the reliability of the evaluation~\cite{seo2023safe}. Other works~\cite{gp_icra,gp_rss,gp_icra2} model traversability assessment using the Gaussian process, enhancing spatial continuity compared to classical elevation maps. Nevertheless, this approach suffers from low computational efficiency and struggles to effectively integrate historical point clouds from LiDAR. 

\subsection{Motion Planning on Uneven Terrain}
\label{subsec:Motion Planning on Uneven Terrain}
In recent years, researchers have extended 2D planning methods to enable car-like robots to navigate uneven terrain autonomously. Krusi et al.~\cite{drivingonpoint} proposed a trajectory planning method based on unordered 3D point clouds obtained from LiDAR, incorporating concurrent terrain assessment into the planning process, but it struggles with real-time performance due to computational limitations. Han et al. \cite{han2023model} proposed a 2D dynamic sampling method for uneven terrain, incorporating a real-time elevation map with physics-based constraints. This approach leverages the model predictive path integral (MPPI) algorithm alongside a low-level controller to enable high-speed robot navigation on challenging terrain. However, dense sampling in complex environments limits its efficiency.

To enhance planning efficiency, Jian et al.~\cite{putn} proposed a hybrid architecture combining plane-fitting, informed-RRT* path searching, Gaussian process regression, and NMPC-based trajectory generation. But its computational scalability is limited by the exponentially increasing complexity of NMPC formulations as the planning horizon extends. To address this issue, Xu et al.~\cite{iros2023} proposed a polynomial-based trajectory optimization framework, incorporating equality constraints to handle the nonholonomic constraints of the robot. While this method improves planning efficiency, the function values corresponding to the equality constraints become very small when the robot starts moving or shifts gears, making it difficult for the optimizer to converge. This not only reduces efficiency but also makes it challenging to adapt their method to scenarios where the robot needs to frequently alternate between forward and backward movements. Inspired by the differential flatness of car-like robots\cite{hzc,unpublished_han}, we find that flat outputs still exist for \textit{terrain-associated kinematics}. Based on this, we propose a new trajectory optimization model which can avoid the introduction of the equational constraint. Besides, we introduce an intermediate variable to eliminate the singularities associated with differential flatness, enabling more efficient planning.

\section{Terrain Pose Mapping}
\label{sec:Terrain Pose Mapping}
In this paper, we use the left superscript to denote the frame in which the vector is located, and ${^B_A}\boldsymbol{R}$ to denote the rotation matrix from frame $A$ to frame $B$. We use $S$ to denote the LiDAR frame, $B$ to denote the robot body frame, and $M$ to denote the map frame. Without labelling, the world frame $W$ is used by default.

On flat terrain, the state of a car-like robot can be represented using $\boldsymbol s_r=[x,y,\theta]^\text T\in SE(2)$. While on uneven terrain, due to the variation in terrain height, we need to represent the state of the robot using a higher dimensional $SE(3)$ state with the position $\boldsymbol{p}_B=[x,y,z]^\text{T}\in \mathbb{R}^3$ and the attitude ${_B}\boldsymbol{R}=[\boldsymbol{x}_b,\boldsymbol{y}_b,\boldsymbol{z}_b]\in SO(3)$.

In our previous work\cite{iros2023}, we proposed \textit{terrain pose mapping} $\mathscr{F}:SE(2)\mapsto\mathbb{R}\times\mathbb{S}^2_+$, where $\mathbb{S}^2_+\triangleq\{\boldsymbol{x}\in\mathbb{R}^3\ |\ \lVert\boldsymbol{x}\rVert_2=1,\boldsymbol{b}_3^\text T\boldsymbol{x}>0\},\boldsymbol{b}_3=[0,0,1]^\text{T}$ to describe the impact
of terrain on the robot. Let $\boldsymbol{x}_{yaw}=[\cos\theta,\sin\theta,0]^\text T$ denote the direction of yaw
angle, the mapping $\mathscr{F}$ can be expressed as two functions $z=f_1(x,y,\theta)$, $\boldsymbol{z}_b=\boldsymbol{f}_2(x,y,\theta)$. Using the Z-X-Y Euler angles to represent the attitude of the
robot, we can obtain:
\begin{align}
&\boldsymbol{p}_B=[x,y,z]^\text{T}=[x,y,f_1(x,y,\theta)]^\text{T},\\
&\boldsymbol{y}_b=\frac{\boldsymbol{f}_2(x,y,\theta)\times\boldsymbol{x}_{yaw}}{\lVert\boldsymbol{f}_2(x,y,\theta)\times\boldsymbol{x}_{yaw}\rVert},\label{eq:xb}\\
&\boldsymbol{x}_b=\boldsymbol{y}_b\times\boldsymbol{f}_2(x,y,\theta)\label{eq:yb}.
\end{align}
Thus, with the help of the mapping $\mathscr{F}$, we can still use
elements in $SE(2)$ to characterize the state of the robot on
uneven terrain. The mapping $\mathscr{F}$ actually implies the height and attitude given by the uneven terrain to $\boldsymbol s_r$.

\begin{figure*}[t]  
		\centering
		{\includegraphics[width=1.0\linewidth]{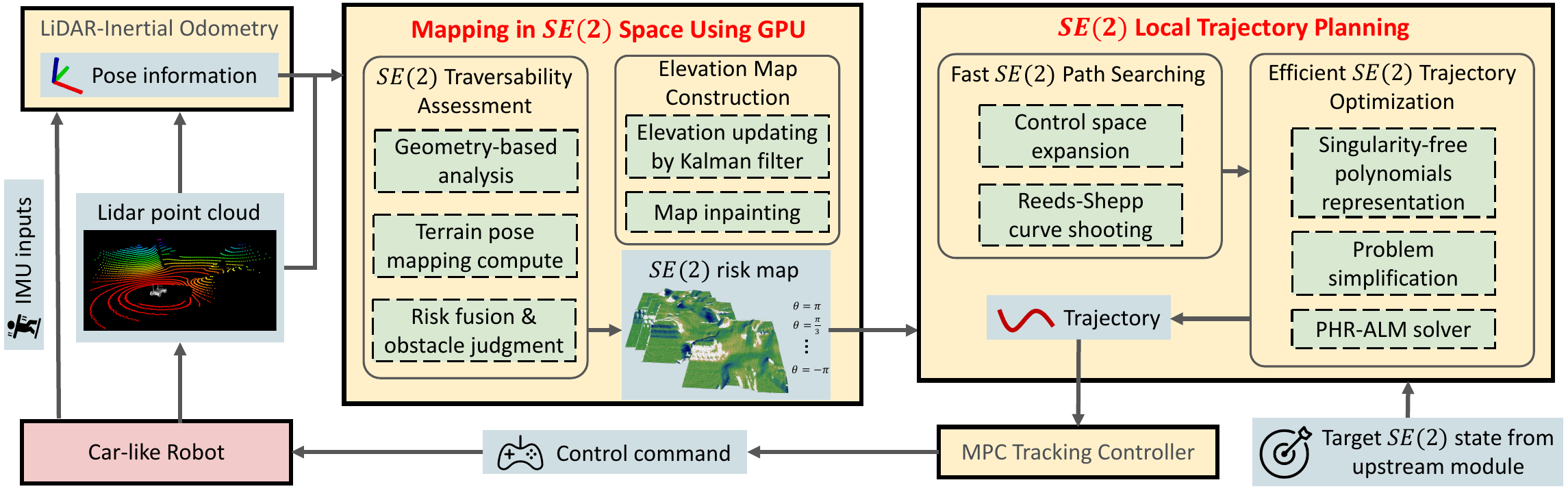}}
		\caption{Overview of the framework. LiDAR-Inertial Odometry (LIO) receives information from IMU and point clouds from LiDAR to calculate pose. Then, the local mapping module acquires the point cloud and poses information to update the elevation and evaluate the traversability in $SE(2)$ space. The resulting risk maps are used by the local planner. After path search and trajectory optimization, the local planner outputs a trajectory in $SE(2)$ space, which is given to the MPC controller for generating final control commands to send to the robot.}
		\label{fig:overview}
\end{figure*}

\section{Overview of the SEB-Naver}
\label{sec:Overview Of The SEB-Naver}
The local navigation problem for car-like robots on uneven terrain can be described as how to obtain a strategy $\pi\sim p(\boldsymbol{u}_t|\boldsymbol{O}_t,\boldsymbol{c}_t)$ such that the robot can successfully reach the destination and satisfy certain conditions during the process, where $\boldsymbol{O}_t$ denotes past observations, $\boldsymbol{c}_t\in SE(2)$ denotes the current target pose, and $\boldsymbol{u}_t$ denotes the control command to send to the robot. The conditions will typically include 1) the risk of robot failure is as small as possible, 2) control commands can be tracked well by the robot, and 3) the robot can move fast while conserving energy. We decompose the problem into several modules, as shown in Fig.~\ref{fig:overview}. 

To handle the first condition, we construct a local map and consider the risk in the local trajectory planning module. In this work, we use the pose information from the localization module and the point clouds from LiDAR as observations $\boldsymbol{O}_t$ to construct a risk map. We mainly consider the second and third conditions in trajectory optimization, which specifically correspond to the kinematic constraints of the robot and the objective function containing smoothness and agility. For these two conditions to be better satisfied, we include an MPC controller downstream of the planner that outputs control commands $\boldsymbol{u}_t$ including steering angle $\delta$ and velocity $v_x$ at a higher frequency. It is only used to track the optimized trajectory without considering obstacle avoidance, risk, etc.

\begin{figure}	
    \centering
    \includegraphics[width=1.0\columnwidth]{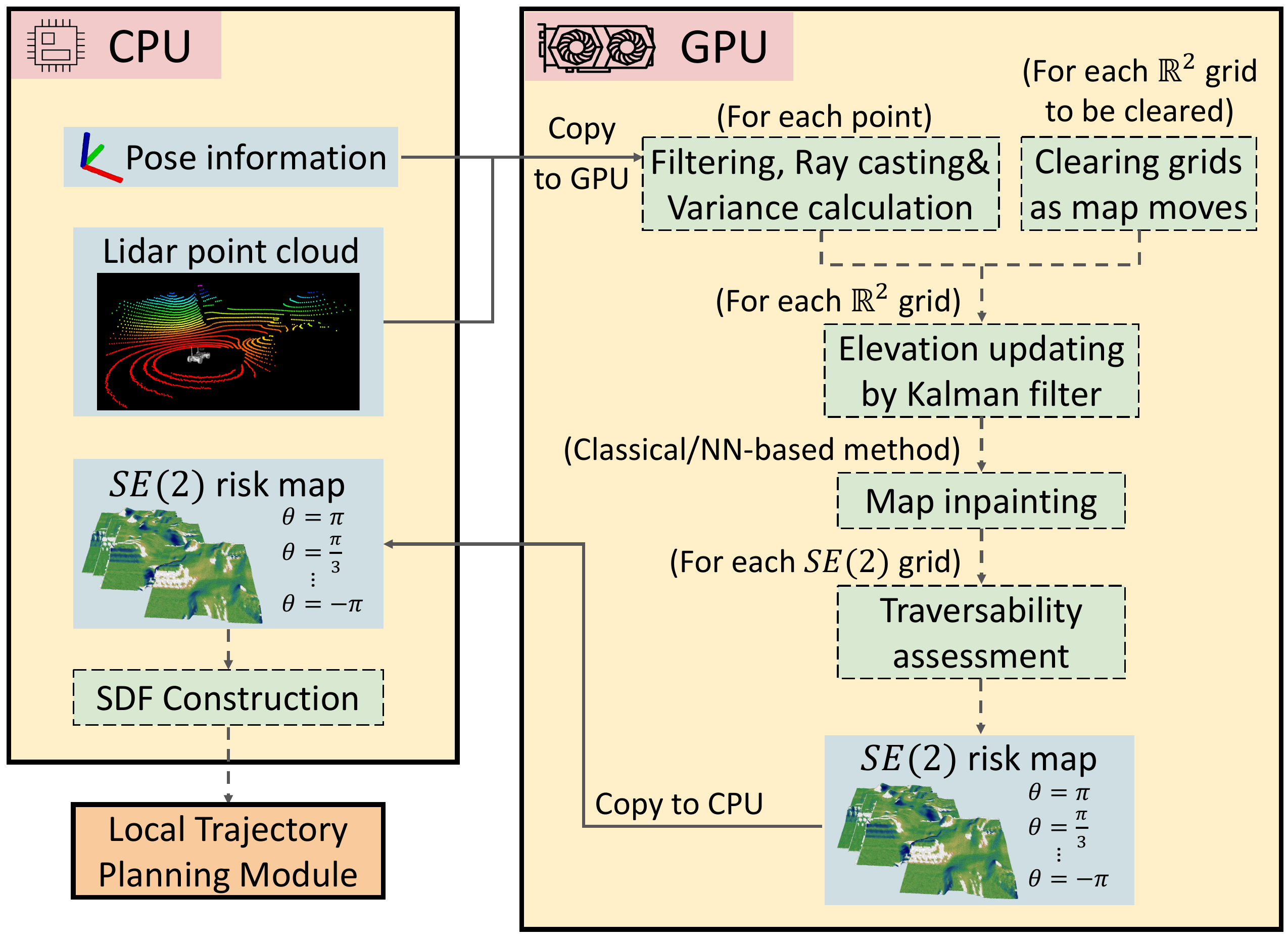}
	\caption{Local mapping process in $SE(2)$ space.}\label{fig:Local Mapping}
\end{figure}

\section{Mapping in SE(2) Space Based on GPU}
\label{sec:Mapping in SE(2) Space Based on GPU}
Fig.~\ref{fig:Local Mapping} illustrates the proposed process of local mapping in $SE(2)$ space. When receiving new pose information and point cloud, the GPU filters the point cloud and computes their variance in parallel. It also resets some elevation grids by ray casting. At the same time, grids outside the range of the map are removed in parallel due to the movement of the map caused by changes in the position of the robot. After these, the elevation map will be updated by Kalman filtering (KF) and can be inpainted by classical methods\cite{fm_inpaint, ns_inpaint} or neural network (NN) based approaches\cite{gan_inpaint, nn_inpaint}. The GPU will compute the traversability for each $SE(2)$ state in parallel using the inpainted elevation map. Eventually, the computed risk map will be sent back to the CPU, and the corresponding signed distance field\cite{sdf} (SDF) will be generated to be used for collision avoidance.

\subsection{Elevation Map Construction}
\label{subsec:Elevation Map Construction}
In this paper, we set $_M\boldsymbol{R} = \boldsymbol{I}$, where $\boldsymbol{I}$ denotes the identity matrix. While the position of the map frame, 
\begin{align}
\boldsymbol{p}_M=l_{\text{res}}\cdot[\text{floor}(x/l_{\text{res}}),\text{floor}(y/l_{\text{res}}),0]^\text T,\label{eq:pm}
\end{align} 
is set to be determined by the robot position, where function $\text{floor}(q)$ serves to take the largest integer not greater than $q$. After acquiring the robot pose and LiDAR point cloud, we first update $\boldsymbol{p}_M$ according to Eq.(\ref{eq:pm}), and then set the state of grids that are no longer within the map to unknown. Besides, we apply parallel ray casting, checking the grid that passes from the LiDAR position to each point, setting the state to unknown if the elevation in the grid is higher than the highest value predicted by the ray.

For points in the point cloud, we ignore the points outside the range of the map. Then, we transform the others to the robot body frame $B$ and filter out the points outside a set height range. The remaining points will be assigned to the corresponding grids depending on the range each grid includes. Let one point be $\boldsymbol{p}_l=[x_l,y_l,z_l]^\text T$. Notice that
\begin{align}
z_l&=\boldsymbol{b}_3^\text{T}\boldsymbol{p}_l=\boldsymbol{b}_3^\text{T}({_S}\boldsymbol{R}^S\boldsymbol{p}_l+\boldsymbol{p}_{S}) \nonumber\\
&=\boldsymbol{b}_3^\text{T}({_B}\boldsymbol{R}({^B_S}\boldsymbol{R}^S\boldsymbol{p}_l+{^B}\boldsymbol{p}_{BS})-\boldsymbol{p}_{B}),
\end{align}
where $\boldsymbol{p}_{S}$ denotes the position of the LiDAR, ${^B}\boldsymbol{p}_{BS}$ denotes the position of the LiDAR in the robot body frame, $^S\boldsymbol{p}_l,{_B}\boldsymbol{R},\boldsymbol{p}_B$ are random variables. We can then compute the Jacobi of $z_l$ as follows:
\begin{align}
\boldsymbol{J}_S&=\frac{\partial z_l}{\partial ^S\boldsymbol{p}_l}=({_B}\boldsymbol{R}{^B_{S}}\boldsymbol{R})^\text{T}\boldsymbol{b}_3,\\
\boldsymbol{J}_R&=\frac{\partial z_l}{\partial{_B}\boldsymbol{R}}=({^B_S}\boldsymbol{R}^S\boldsymbol{p}_l+{^B}\boldsymbol{p}_{BS})^{\wedge}{_B}\boldsymbol{R}^\text T\boldsymbol{b}_3,\\
\boldsymbol{J}_{B}&=\frac{\partial z_l}{\partial \boldsymbol{p}_B}=-\boldsymbol{b}_3,
\end{align}
where the operations $\boldsymbol{q}^{\wedge}$ takes the vector $\boldsymbol{q}\in\mathbb{R}^3$ to form a skew-symmetric matrix. We can obtain the variance of $z_l$:
\begin{align}
\sigma^2_{z_{l}}&=\boldsymbol{J}_S^\text{T}\Sigma_S\boldsymbol{J}_S+\boldsymbol{J}_R^\text{T}\Sigma_R\boldsymbol{J}_R+\boldsymbol{J}_B^\text{T}\Sigma_B\boldsymbol{J}_B,
\end{align}
where $\Sigma_S, \Sigma_R, \Sigma_B$ are covariance matrices. $\Sigma_S$ is determined by the sensor model, while $\Sigma_R, \Sigma_B$ are given by the localization module.

After calculating the variances for all points, we update the elevations of the $\mathbb{R}^2$ grids in parallel using a one-dimensional Kalman filter and handle multiple points that fall into the same grids by Mahalanobis distance-based method, in the same way as the classical elevation map\cite{fankhauser2018probabilistic}.



\subsection{Traversability Assessment}
\label{subsec:Traversability Assessment}
\begin{algorithm}[t]
    \caption{Traversability and \textit{terrain pose mapping} assessment in state $\boldsymbol{s}_r\in SE(2)$}
    \label{alg:traversability assessment}
    \KwIn{elevation map $\mathcal{M}$, state $\boldsymbol{s}_r\in SE(2)$, parameters of ellipse $(e_x,e_y)$, weights $\boldsymbol{w}_r\in\mathbb{R}^3$, Constants $\kappa_{\text{max}},\varphi_{x_{\text{max}}},\varphi_{y_{\text{max}}}$}
    \KwOut{$\text{Risk}(\boldsymbol{s}_r), \boldsymbol{z}_b(\boldsymbol{s}_r)$}
    \Begin
    {
        $\mathcal{P}\leftarrow\text{FindEllipticalPoints}(\mathcal{M},\boldsymbol{s}_r,e_x,e_y)$\;
        $\boldsymbol{p}_{\text{mean}}\leftarrow\text{GetMeanPosition}(\mathcal{P})$\;
        $Cov\leftarrow\text{ZeroMatrix3}()$\;
        \For{\textbf{each} $\boldsymbol{p}_j \in\mathcal{P} $}
        {
            $\boldsymbol{p}_{e}\leftarrow \boldsymbol{p}_{j}-\boldsymbol{p}_{\text{mean}}$\;
            $Cov\leftarrow Cov+\boldsymbol{p}_e\boldsymbol{p}_e^\text{T}$\;
        }
        $Cov\leftarrow Cov/\text{NumOf}(\mathcal{P})$\;
        $\boldsymbol{z}_{b},\kappa_{\text{ter}}\leftarrow\text{GetMinEigenVecWithCurv}(Cov)$\;
        \If{$\kappa_{\text{ter}}>\kappa_\text{max}$}
        {
            \textbf{return} $(1,\boldsymbol{z}_{b})$.
        }
        $\boldsymbol{x}_{b},\boldsymbol{y}_{b}\leftarrow\text{GetXbYb}(\boldsymbol{z}_{b}, \boldsymbol{s}_r)$\;
        $\varphi_{x}\leftarrow\lvert\arcsin(\boldsymbol{b}_{3}^\text T\boldsymbol{x}_{b})\rvert,\varphi_{y}\leftarrow\lvert\arcsin(\boldsymbol{b}_3^\text T\boldsymbol{y}_{b})\rvert$\;
        \If{$\varphi_{x}>\varphi_{x_{\text{max}}}\vee\varphi_{y}>\varphi_{y_{\text{max}}}$}
        {
            \textbf{return} $(1,\boldsymbol{z}_b)$.
        }
        $\boldsymbol{r}\leftarrow\left[\frac{\kappa_{\text{ter}}}{\kappa_{\text{max}}},\frac{\varphi_{x}}{\varphi_{x_{\text{max}}}},\frac{\varphi_{y}}{\varphi_{y_{\text{max}}}}\right]^\text T$\;
        $\text{Risk}(\boldsymbol{s}_r)\leftarrow\boldsymbol{r}^\text T\boldsymbol w_r$\;
        \textbf{return} $(\text{Risk}(\boldsymbol{s}_{r}), \boldsymbol{z}_{b})$.
    }
\end{algorithm}
We propose Algorithm \ref{alg:traversability assessment} to evaluate the risk and \textit{terrain pose mapping} in state $\boldsymbol{s}_r$. Similar to our previous work\cite{iros2023}, we take the points corresponding to the elevation map in an elliptical area related to the robot's size, obtaining $\boldsymbol{z}_b$ and the approximate terrain curvature $\kappa_{\text{ter}}$ by geometry-based analysis. After using the resultant $\boldsymbol{z}_b$ to calculate $\boldsymbol{y}_b$ and $\boldsymbol{x}_b$ by Eq.(\ref{eq:xb}) and Eq.(\ref{eq:yb}), we check the angle of these two vectors with the horizontal plane to determine the traversability, respectively. We fuse the curvature $\kappa_{\text{ter}}$, representing the geometric information of the terrain, and the angles $\varphi_x$ and $\varphi_y$, representing the attitude of the robot, by weighted summation to obtain the final risk. Regions with $\text{Risk}(\boldsymbol s_r)=1$ are judged as explicit obstacles.

\section{Local Trajectory Planning}
\label{sec:Local Trajectory Planning}
\subsection{Trajectory Parameterization}
\label{subsec:Trajectory Parameterization}
The kinematic equations for car-like robot driving on uneven terrain can be represented as follows:
\begin{align}
&\dot{\boldsymbol{p}}_B=v_x\cdot\boldsymbol{x}_b,\label{eq:step}\\
&{_B}\dot{\boldsymbol{R}}={_B}\boldsymbol{R}\lfloor \frac{v_x\tan \delta}{L_w}\cdot\boldsymbol{z}_b \rfloor,\label{eq:ster}
\end{align}
where $L_\text{w}$ is the wheelbase length of the robot, the operation $\lfloor*\rfloor$ takes a vector to a skew-symmetric matrix. In our previous work, we demonstrated the use of \textit{terrain pose mapping} to reduce the dimensionality of the robot to $SE(2)$ space, as well as the efficiency of trajectory planning brought by polynomial parameterizations. To satisfy the nonholonomic constraints of the robots, it introduces the following equational constraint in the trajectory optimization modelling:
\begin{align}
\dot x(t)\sin\theta(t)-\dot y(t)\cos\theta(t)=0.\label{con:nonh}
\end{align}
However, given that $v_x(t) = \eta(t)\sqrt{\dot x^2(t)+\dot y^2(t)}$, the left-hand side of Eq.(\ref{con:nonh}) becomes very small when the robot's speed is low (e.g., when the robot starts moving or changes gears), where $\eta(t)=1\vee-1$ denotes the robot is moving forward or backward. This leads to numerical issues in the optimization process, which not only reduces efficiency but also makes it difficult to extend the method to scenarios where the robot needs to frequently alternate between moving forward and backward. To address this problem, we draw inspiration from the differential flatness of car-like robots\cite{hzc,unpublished_han}, introducing an intermediate variable $s(t)$ to represent the trajectory. 

Reconsidering Eq.(\ref{con:nonh}), if there is $\dot x^2(t)+\dot y^2(t)\neq0$, we can explicitly compute $\theta(t)=\text{arctan2}(\eta(t)\dot y(t),\eta(t)\dot x(t))$. So we can rewrite Eq.(\ref{con:nonh}) with $s(t)$ as:
\begin{align}
\dot x(s)\dot s(t)\sin\theta(t)+\dot y(s)\dot s(t)\cos\theta(t)=0.
\end{align}
Thus, we can add the constraint $\dot x^2(s)+\dot y^2(s)>\delta_+$, obtaining $\theta(t)=\text{arctan2}(\eta(t)\dot y(s),\eta(t)\dot x(s))$, eliminating the equation constraint (\ref{con:nonh}), where $\delta_+>0$ is a constant. This approach avoids the singularity of the function $\text{arctan2}(*)$ in Eq.(\ref{con:nonh}) when $\dot x^2(t)+\dot y^2(t)=0$.

Similar to the work\cite{iros2023}, we use quintic piecewise polynomials with four times continuously differentiable at the segmented points to represent state trajectories. Each piece of trajectory can be denoted as:
\begin{align}
&x_i(s)=\boldsymbol{c}_{x_i}^\text{T}\boldsymbol{\beta}(s),\quad s\in[0,s(T_p) - s(0)],\\
&y_i(s)=\boldsymbol{c}_{y_i}^\text{T}\boldsymbol{\beta}(s),\quad s\in[0,s(T_p) - s(0)],\\
&s_i(t)=\boldsymbol{c}_{s_i}^\text{T}\boldsymbol{\beta}(t),\quad\  t\in[0,T_p],
\end{align}
where $i=1,2,...,\text N$ is the index of piecewise polynomial, $T_p$ is the duration of one piece of the trajectory, $\boldsymbol{c}_*\in\mathbb{R}^6,*=\{x_i,y_j,s_i\}$ is the coefficient of polynomial, $\boldsymbol{\beta}(t)=[1,t,t^2,...,t^5]^\text{T}$ is the natural base. 

Using the \textit{terrain pose mapping} and the state transfer equations Eq.(\ref{eq:step}) and (\ref{eq:ster}), letting $\boldsymbol{y}_{yaw}=[-\sin\theta,\cos\theta,0]^\text{T}$, we can derive the control inputs and some kinematic variables from the parameterized trajectory as follows:
\begin{align}
v_x &= \frac{\dot s(t) \sqrt{\dot x^2(s) + \dot y ^2(s)}}{\boldsymbol{x}_b^\text T\boldsymbol{x}_{yaw}},\\
\omega_z &= \frac{\dot x(s) \ddot y(s) - \dot y(s)\ddot x(s)}{(\dot x^2(s) + \dot y^2(s)) \boldsymbol z_b^\text T \boldsymbol b_3} \cdot \dot s(t)\\
\delta&=\text{arctan2}(L_w\omega_z,v_x)\\
a_x&= \frac{ \dot s^2(t) \boldsymbol v_{ds}^\text T\boldsymbol a_{ds} + \ddot s(t)\boldsymbol v_{ds}^\text T\boldsymbol v_{ds}}{\sqrt{\dot x^2(s) + \dot y^2(s)} \boldsymbol x_b^\text T \boldsymbol x_{yaw}}+ g \cdot\boldsymbol x_b^\text T\boldsymbol b_3,\\
a_y&=\frac{\dot x(s) \ddot y(s)  - \dot y(s)\ddot x(s)}{\sqrt{\dot x^2(s) + \dot y^2(s)} \boldsymbol y_b^\text T \boldsymbol y_{yaw}} \dot s^2(t)+g\cdot\boldsymbol{y}_b^\text{T}\boldsymbol{b}_{3},
\end{align}
where $\boldsymbol{v}_{ds}=[\dot x(s), \dot y(s)]^\text T$, $\boldsymbol{a}_{ds} = [\ddot x(s), \ddot y(s)]^\text T$. $a_x,a_y$ denote longitude and latitude acceleration, respectively.

\subsection{Optimization Problem}
\label{subsec:Optimization Problem}
In this paper, we formulate the local trajectory optimization problem for Car-like robots on uneven terrain as:
\begin{align}
\min_{\boldsymbol{c},\boldsymbol{e}_m,T_{f}}&f(\boldsymbol{c},\boldsymbol{e}_m,T_{f})=\int_0^{T_f}\boldsymbol{j}(t)^\text{T}\boldsymbol{j}(t)dt+\rho_tT_f\nonumber\\
&\quad\quad\quad\quad\quad\quad+\rho_r\int_0^{T_f}\text{Risk}^2(\boldsymbol{s}_r(t))dt\label{problem:opt}\\
&s.t.\ \ \boldsymbol{M}_{p}(\boldsymbol{S})\boldsymbol{c}_{p}=\boldsymbol{b}_{p}(\boldsymbol{P},\boldsymbol{e}_m),\label{con:mincop}\\
&\quad\quad \boldsymbol{M}_{s}(T_f)\boldsymbol{c}_{s}=\boldsymbol{b}_{s}(\boldsymbol{S}),\label{con:mincos}\\
&\quad\quad T_f>0,\quad \dot{x}^2(s)+\dot y^2(s)\ge\delta_+,\label{con:slackvf}\\
&\quad \quad \delta^2(t)\le\delta_\text{max} ,\quad v_x^2(t)\leq v_\text{mlon}^2,\label{con:dyn_begin}\\
&\quad \quad a_x^2(t)\leq a_\text{mlon}^2,\quad a_y^2(t)\leq a_\text{mlat}^2,\label{con:dyn_end}\\
&\quad \quad \varphi_x^2(t)\le\varphi^2_{x_\text{max}},\quad \varphi^2_y(t)\le\varphi^2_{x_\text{max}},\label{con:attitude}\\
&\quad \quad \text{Risk}(\boldsymbol{s}_r(t))\le r_\text{max},\quad\boldsymbol{\mathscr{G}}_c(\boldsymbol{s}_r(t))\geq d_\text{min},\label{con:safe}
\end{align}
where $\boldsymbol{c}_p=[[\boldsymbol{c}_{x_1},\boldsymbol{c}_{y_1}]^\text T,[\boldsymbol{c}_{x_2},\boldsymbol{c}_{y_2}]^\text T,...,[\boldsymbol{c}_{x_\text N},\boldsymbol{c}_{y_\text N}]^\text T]^\text{T}\in\mathbb{R}^{6\text N\times2}$,$\boldsymbol{c}_{s}=[\boldsymbol{c}_{{s}_1}^\text T,\boldsymbol{c}_{{s}_2}^\text T,...,\boldsymbol{c}_{{s}_\text N}^\text T]^\text T\in\mathbb{R}^{6\text N\times1}$ are coefficient matrices. $\boldsymbol{c}=\{\boldsymbol{c}_p,\boldsymbol{c}_{s}\}$. For the other optimization variables, $\boldsymbol{e}_m\in\mathbb{R}^{4\times\Omega}$, where each column of it is a boundary variable $\boldsymbol{e}_m^w=[x(s_w),y(s_w),\dot x(s_w),\dot y(s_w)],w=1,2,...,\Omega$, denoting the moment point at the $w^\text{th}$ gear switch. $T_f$ denote the total duration of the trajectory. In the objective function (\ref{problem:opt}), $\boldsymbol{j}(t)=[x^{(3)}(t),y^{(3)}(t)]^\text T$ denote the jerk of the trajectories. $\rho_t>0,\rho_r>0$ are constants to give the trajectory some aggressiveness and reduce risk, respectively. Eq.(\ref{con:mincop}) and Eq.(\ref{con:mincos}) denote the combinations of the continuity constraints for polynomials and boundary conditions of the trajectory.

$\boldsymbol{P}\in\mathbb{R}^{2\times(\text N-1)}$ are the segment points of the trajectories, $\boldsymbol{S}\in\mathbb{R}^{\text N\times1}$ denote the value of each segment of the intermediate variable. Thus $\boldsymbol{M}_{p},\boldsymbol{M}_{s}\in\mathbb{R}^{(6\text N-2)\times6\text N}$. In practice, we set $\dot x^2(0)+\dot y^2(0)=1$ and $\delta_+=0.9$ for optimization.

Conditions (\ref{con:slackvf}) denote the positive duration and the constraints brought by the introduction of $s(t)$ mentioned in the last subsection. Conditions (\ref{con:dyn_begin}) and (\ref{con:dyn_end}) are dynamic feasibility constraints, including limitation of steering angle $\delta(t)$, velocity $v_x(t)$, longitude acceleration $a_x(t)$, and latitude acceleration $a_y(t)$, where $\delta_\text{max},v_\text{mlon},a_\text{mlon},a_\text{mlat}$ are constants. Conditions (\ref{con:attitude}) are used to ensure that the robot's attitude remains within a safe range.

Conditions (\ref{con:safe}) represent safety constraints, ensuring the risk stays below the threshold $r_\text{max}$ and avoiding explicit obstacles. To enforce this, we use the SDF $\mathscr{G}_c(\boldsymbol{s}_r):SE(2)\mapsto\mathbb{R}$, where value at each state is the distance to the edge of the nearest region, with negative values inside obstacles. $d_\text{min}$ defines the obstacle avoidance threshold.

\subsection{Problem Solving}
\label{subsec:Problem Solving}
To address the equality constraints (\ref{con:mincop}) and (\ref{con:mincos}), we refer to the method proposed in work\cite{minco}, adding the constraints of the trajectories having zero second-order derivatives at the start, termination and gear switch time to ensure that the matrices $\boldsymbol{M}_p,\boldsymbol{M}_\mathfrak{s}$ are invertible. This allows us to transform the optimization variables into $\{\boldsymbol{P},\boldsymbol{S},\boldsymbol{e}_m,T_f\}$, while simultaneously reducing the dimensionality of the optimization variables.

To deal with the constraint $T_f>0$, we convert the optimization variable $T_f$ to $\tau$ using the  logarithmic function: $\tau=\ln (T_f)$. For the remaining inequality constraints, we discretize each duration $T_{p}=T_f/\text{N}$ as $\text K$ time stamps $\tilde{t}_{p}=(p/\text K)\cdot T_{p},p=0,1,...,\text K-1$, and impose the constraints at these timestamps. We also discretize $\text{Risk}(\boldsymbol{s}_r(t))$ in this way and accumulate it to obtain its integral. For conditions (\ref{con:safe}), we adopt trilinear interpolation to obtain the values and gradients of the function $\mathscr{G}_c$ and $\text{Risk}$, where operations on manifold\cite{manifold_cpt} are applied for the handling of $SO(2)$ space.

Then, the Powell-Hestenes-Rockafellar augmented Lagrangian method\cite{alm} (PHR-ALM), an effective approach for solving large-scale problems and problems with complex constraints, is utilized to address the simplified trajectory optimization problem. To obtain the initial solution, we employ a lightweight hybrid-A* algorithm to perform path planning on a $SE(2)$ grid map. The Reed-Shepp curve\cite{rscurve} is used to shoot toward the goal state, allowing the search process to terminate early.

\begin{figure}[t]
    \centering
    \includegraphics[width=1.0\columnwidth]{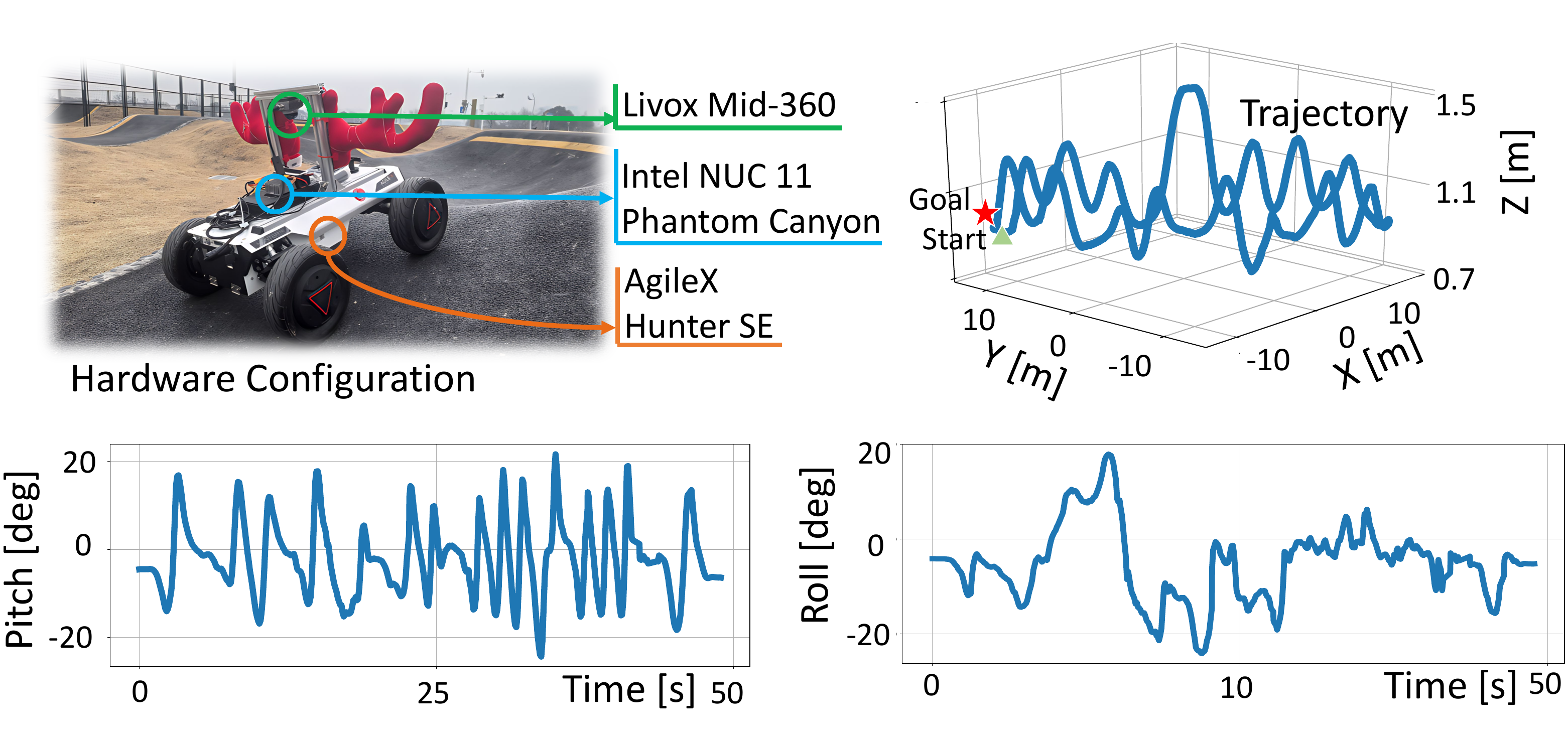}
    \caption{One test of the SEB-Naver on the pump track.}
    \label{fig:real_car}
    \vspace{-0.4cm}
\end{figure}

\section{Experiments}
\label{sec:Experiments}
\subsection{Implementation Details}
We adopt CUDA\cite{cuda} to implement the proposed real-time traversability assessment pipeline for $SE(2)$ grids, further deploying SEB-Naver on a car-like robot for real-world experiments, as shown in Fig.~\ref{fig:real_car}. FAST-LIO2\cite{fast-lio2} is used for localization. We also build simulation environments based on EPFL terrain generator\footnote{https://github.com/droduit/procedural-terrain-generation} to conduct comparative experiments with existing methods. All of them are run on a desktop with an Intel i5-14600 CPU and Nvidia GeForce RTX1660 GPU.

\subsection{Comparison Experiments}
\label{subsec:Comparison Experiments}
\textit{1) Mapping in $SE(2)$:} 
We compared the computation time of the proposed local mapping method with the baseline (CPU) \cite{iros2023} on two devices: the above-mentioned desktop computer and an edge computing device, Jetson Xavier NX. To analyze the computation time under different grid resolutions, we set the dimensions of the $\mathbb{R}^2$ space to range from $8m\times8m$ to $18m\times18m$ with a resolution of $0.1m\times0.1m$, and the number of grids in the $SO(2)$ space varied between $8$ and $32$. For the proposed method, the processing includes all features shown in Fig.~\ref{fig:Local Mapping}: point cloud filtering, ray casting, variance computing, height update and traversability assessment. The baseline method has the same pipeline except for ray casting. Both methods utilize nearest-neighbor interpolation for elevation inpainting.

\begin{figure}[t]
    \centering
    \includegraphics[width=1.0\columnwidth]{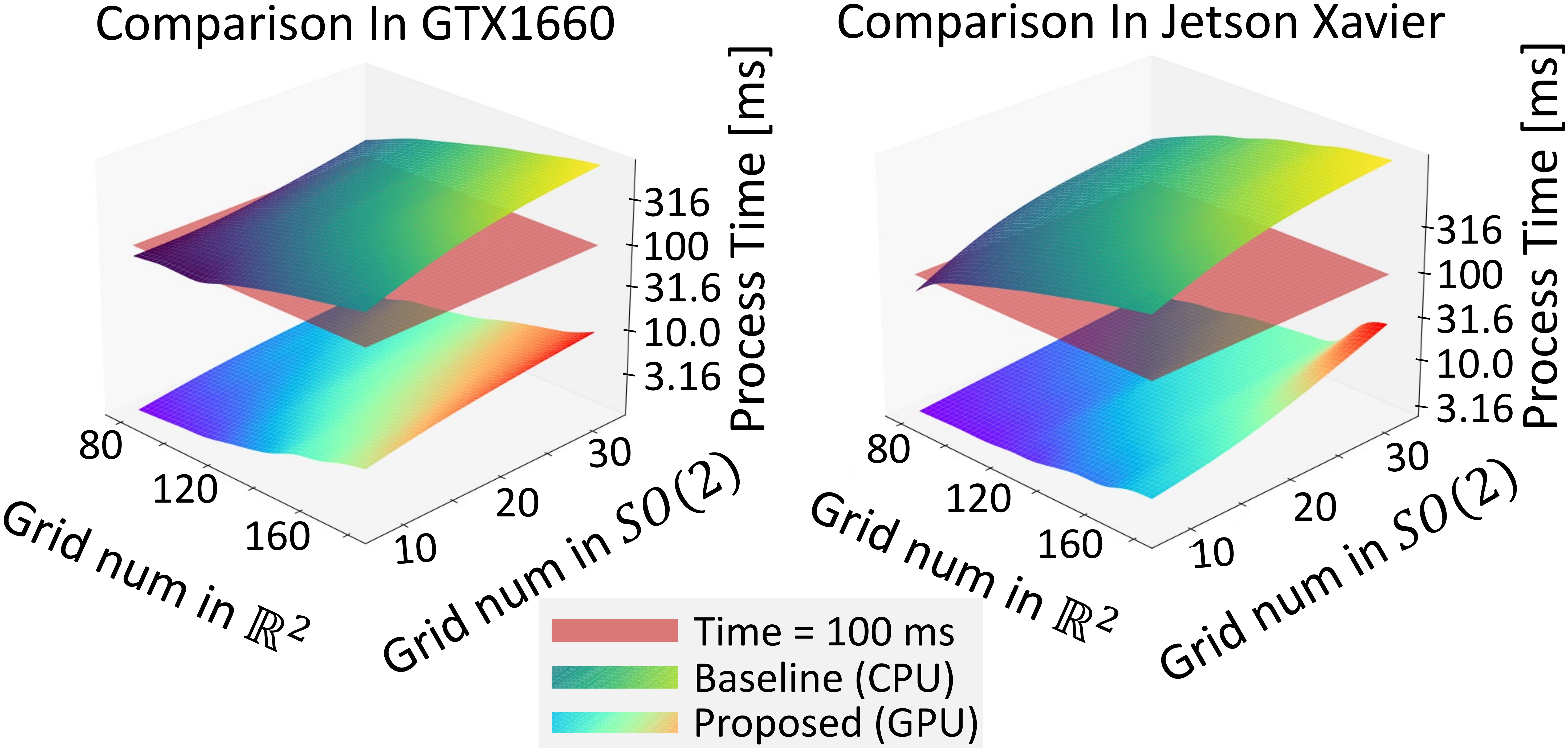}
    \caption{Processing time comparison of $SE(2)$ grids.}
    \label{fig:map process}
\end{figure}

The results are presented in Fig.~\ref{fig:map process}. Since our processing is done on GPU, the processing time remained short even for large numbers of grids. Specifically, our method processes the densest grid configurations of 972,000 grids within $50$ milliseconds, meeting the real-time requirement set by the frequency of 20 Hz. On the other hand, the baseline method requires over $100$ milliseconds in most cases. This shows that our mapping approach is more efficient than the baseline in handling grids in $SE(2)$ space despite the additional ray-casting computations performed in our pipeline.

\begin{figure}[t]
    \centering
    \includegraphics[width=1.0\columnwidth]{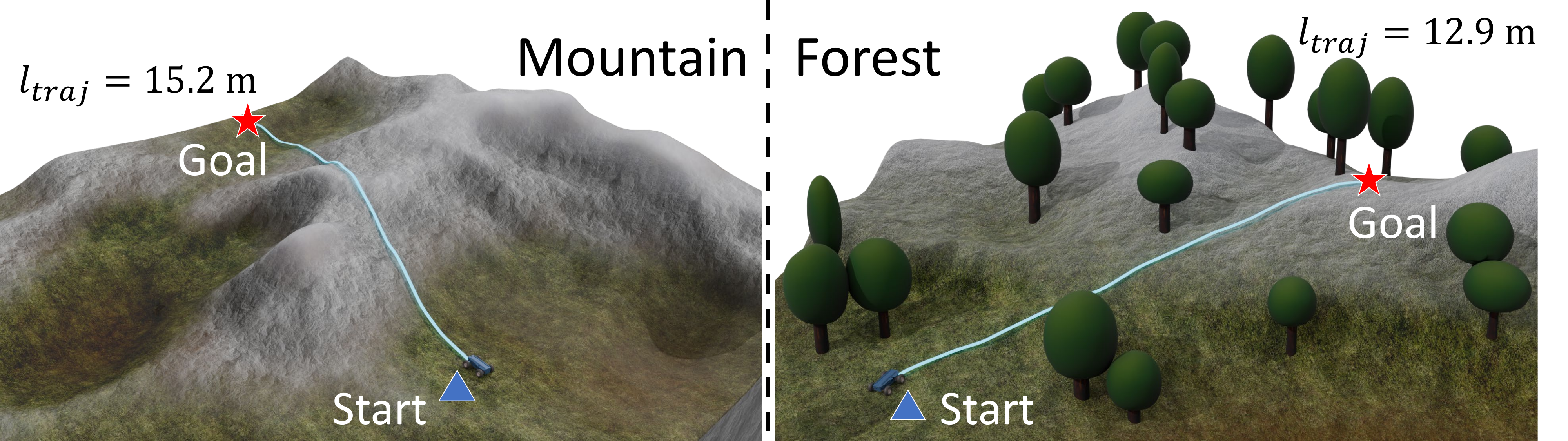}
    \caption{Two examples of benchmark scenarios. $l_{traj}$ denotes the length of the trajectory generated by SEB-Naver.}
    \label{fig:scenarios}
    \vspace{-0.4cm}
\end{figure}

\begin{table}[t]
	\small
	\centering
	\renewcommand\arraystretch{1.2}
	\caption{\label{tab:table1} Methods Qualitative Comparison }
	\begin{tabular}{c|cccc}
		\hline
		\multirow{2}{*}{Method}& \multirow{2}{*}{Reversing} & \multicolumn{1}{c}{Acceleration} & \multirow{2}{*}{Continuity} \\ 
	    &	& \multicolumn{1}{c}{information}
	    
		\\ \hline
		\multirow{1}{*}{\begin{tabular}[c]{@{}c@{}}Proposed\end{tabular}}
		& \textbf{Support}                
		& \textcolor[RGB]{61,145,64}{\ding{52}}    &      \textbf{ 4-order }  \\ \hline
		\multirow{1}{*}{\begin{tabular}[c]{@{}c@{}}Xu's\cite{iros2023}\end{tabular}}
		& No support 	&\textcolor[RGB]{61,145,64}{\ding{52}} &\textbf{4-order} 
		\\ \hline
		\multirow{1}{*}{\begin{tabular}[c]{@{}c@{}}Jian's\cite{putn} \end{tabular}} 
		& \textbf{Support} &\textcolor[RGB]{227,23,13}{\ding{56}}	& 1-order  \\ \hline
	\end{tabular}
    \vspace{-0.2cm}
\end{table}

\begin{figure}[t] 
    \centering
    \includegraphics[width=1.0\columnwidth]{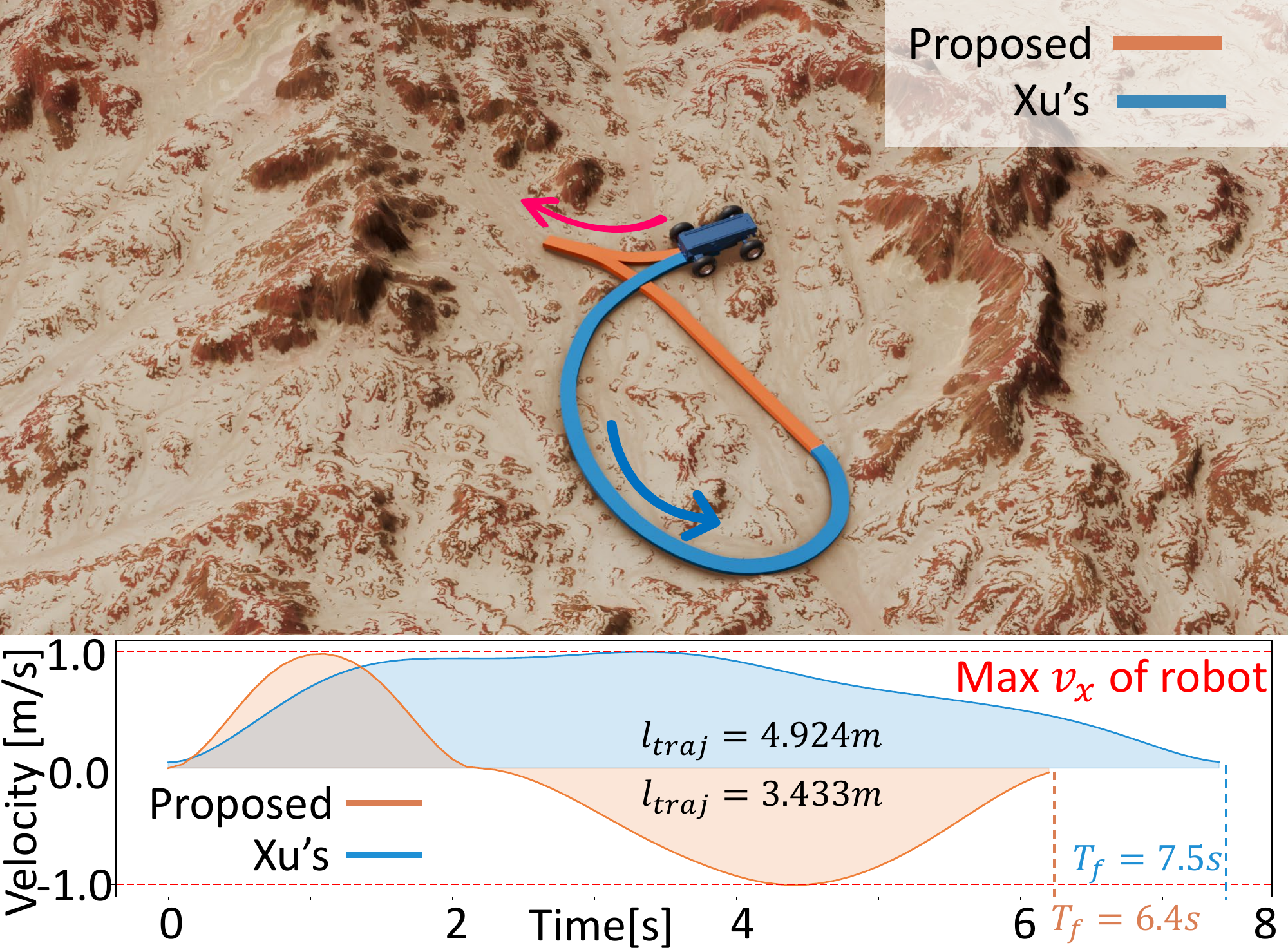}
    \caption{Comparison of velocity curves on a snowy mountain.}
    \label{fig:reverse}
    \vspace{-0.4cm}
\end{figure}

\textit{2) Trajectory planning:} 
We compared the proposed planner with Xu's\cite{iros2023} and Jian's\cite{putn} methods in the terrains generated by the terrain generator mentioned before. Fig.~\ref{fig:scenarios} shows two examples of uneven mountainous and forested. The steering angle, velocity, longitude acceleration, and latitude acceleration of the car-like robot are limited to $\delta_\text{max}=0.785\text{ rad}$, $v_\text{mlon}=1.0m/s$, $a_\text{mlon}=5.0m/s^2$,$a_\text{mlat}=10.0m/s^2$, respectively. The attitude limitations of the robot are set to $\varphi_{x_\text{max}}=\varphi_{y_\text{max}}=0.52\text{ rad}$. We first conducted a qualitative comparison, as shown in Table~\ref{tab:table1}. Since both the proposed method and Xu's\cite{iros2023} method use continuous polynomials rather than discrete states for trajectory parameterization, they exhibit better continuity and can provide higher-order acceleration information for the robot. To evaluate the impact of supporting reversing, we further compared the velocity curve with Xu's\cite{iros2023} method on an uneven snowy mountain terrain. As shown in Fig.~\ref{fig:reverse}, our approach expands the solution space by considering the possibility of negative longitudinal velocities of the robot, which results in shorter trajectory duration and length. This is a huge advantage when robots are navigating on complex, uneven terrain.

\begin{table}[t]
    \small
    \centering
    \renewcommand\arraystretch{1.2}
    \caption{Benchmark comparison of trajectory planner} 
    \label{tab:bk}
    \begin{tabular}{l|l|ccc}
\hline
\multicolumn{1}{c|}{\multirow{2}{*}{Scenario}} & \multicolumn{1}{c|}{\multirow{2}{*}{Method}} & Avg. $t_p$     & Avg. $T_f$     & Avg. $l_{traj}$               \\
\multicolumn{1}{c|}{}                          & \multicolumn{1}{c|}{}                        & (ms)           & (s)            & (m)                           \\ \hline
\multirow{3}{*}{Mountain}                  & Proposed                                      & \textbf{49.30} & \textbf{11.25} & \underline{8.13} \\
                                               & Xu's\cite{iros2023}         & 142.39         & 13.88          & 11.64                         \\
                                               & Jian's\cite{putn}           & 224.50         & 15.14          & 6.77                          \\ \hline
\multirow{3}{*}{Forest}                        & Proposed                                      & \textbf{57.42} & \textbf{12.23} & \underline{8.63} \\
                                               & Xu's\cite{iros2023}         & 130.74         & 15.13          & 12.70                         \\
                                               & Jian's\cite{putn}           & 289.22         & 16.65          & 7.22                          \\ \hline
\end{tabular}
    \vspace{-0.6cm}

\end{table}

We also conducted benchmark comparisons in $20$ simulated environments similar to those in Fig.~\ref{fig:scenarios}. A total of two hundred starts and goals are sampled in each of these scenarios. The results are shown in Table~\ref{tab:bk}, where $t_p$ represents planning time, $T_f$ represents trajectory duration, and $l_{traj}$ represents the trajectory length. The results indicate that, by leveraging differential flatness to simplify the state representation of the car-like robot on uneven terrain, our method achieves several-fold improvements in computational efficiency compared to Jian's\cite{putn} and Xu's\cite{iros2023} methods. Although the average length of the trajectory generated by our method is longer than that of Jian's\cite{putn}, the trajectories generated by Jian's\cite{putn} cannot be tracked fast by the robot due to the lack of consideration for trajectory duration in its optimization model. While Xu’s\cite{iros2023} also incorporates $T_f$ as an optimization variable, its inability to support reversing limits its solution space, excluding better solutions.

\subsection{Real-world Experiments }
\label{subsec:Real-World Experiments}
Real-world experiments are conducted on various types of uneven terrain, including underground parking lots, grasslands, forests, farmland, and pump track, as shown in Fig.~\ref{fig:real_exp}. These autonomous navigation tests demonstrate the effectiveness and practicality of SEB-Naver in complex real-world environments. Fig.~\ref{fig:real_car} illustrates the changes in position and attitude of the robot on a pump track. More demonstrations can be found in the attached multimedia. 

\section{Conclusion \& Limitations}
\label{sec:Conclusion & Limitations}
In this paper, we proposed SEB-Naver, a novel $SE(2)$-based local navigation framework for car-like robots on uneven terrain, which leverages GPU to allow real-time traversability assessment for $SE(2)$ grids, introducing differential flatness for efficient trajectory optimization. Extensive simulations and real-world experiments validate its effectiveness for uneven terrain navigation. The limitation of SEB Naver lies in its reliance on accurate terrain modeling and state estimation, which may be unreliable in dynamic environments or scenarios with feature degradation. Additionally, it does not yet handle complex dynamics like slip or deformable terrain. Future work will focus on addressing these issues to improve robustness and applicability.

\newlength{\bibitemsep}\setlength{\bibitemsep}{0.00\baselineskip}
\newlength{\bibparskip}\setlength{\bibparskip}{0pt}
\let\oldthebibliography\thebibliography
\renewcommand\thebibliography[1]{
    \oldthebibliography{#1}
    \setlength{\parskip}{\bibitemsep}
    \setlength{\itemsep}{\bibparskip}
}
\bibliography{references}

\end{document}